\documentclass{article}
\usepackage{bbm}
\usepackage[table]{xcolor}
\usepackage{multirow}
\usepackage{svg}

\usepackage[english]{babel}

\usepackage[letterpaper,top=2cm,bottom=2cm,left=3cm,right=3cm,marginparwidth=1.75cm]{geometry}

\usepackage{amsmath}
\usepackage{graphicx}
\usepackage[colorlinks=true, allcolors=blue]{hyperref}
\usepackage{booktabs} % To thicken table lines
\usepackage{caption}

\usepackage{mathtools,rotating,amsmath,amssymb}
\usepackage{amsthm}
\usepackage{graphicx}
\usepackage{bbm}
\usepackage{bigstrut,morefloats}

\newcommand{\eat}[1]{}

\newcommand{\LMNets}{Inference Systems}
\newcommand{\LMNet}{Inference System}
\newcommand{\dataitem}{query}
\newcommand{\Dataitem}{Query}

\newcommand{\FVISFull}{Filter-Vote}
\newcommand{\VISFull}{Vote}

\usepackage[linesnumbered,ruled,vlined]{algorithm2e}

\newtheorem{theorem}{Theorem}

\newtheorem{lemma}[theorem]{Lemma}
\newtheorem{definition}{Definition}

\usepackage{subcaption}

\usepackage{multirow}

\usepackage[most]{tcolorbox} % loads the tcolorbox package with most common features

\newtcolorbox[auto counter, number within=section]{mybox}[2][]{
    colback=blue!5!white,    % background color
    colframe=blue!75!black,  % frame color
    fonttitle=\bfseries,     % title font
    title=Main Contributions, % box title
    #1                      % allow for additional options
}

\usepackage{mathrsfs}

\usepackage{wrapfig}

\newif\ifcomments
\commentstrue
\ifcomments
    \providecommand{\ion}[1]{{\color{blue}{[ion: #1]}}}
 \else
    \providecommand{\ion}[1]{}
\fi

\title{Are More LM Calls All You Need? \\Towards the Scaling Properties of Compound AI Systems
}
\author{Lingjiao Chen$^\dag$, Jared Quincy Davis$^\dag$, Boris Hanin$^\S$, \\
Peter Bailis$^\dag$, Ion Stoica$^\ddag$,  Matei Zaharia$^\ddag$, James Zou$^\dag$\\\\
$^\dag$Stanford University, $^\ddag$UC Berkeley,  $^\S$Princeton University
}
\date{}

\begin{document}
\maketitle

\begin{abstract}
Many recent state-of-the-art results in language tasks were achieved using compound systems that perform multiple Language Model (LM) calls and aggregate their responses. However, there is little understanding of how the number of LM calls -- \emph{e.g.,} when asking the LM to answer each question multiple times and taking a majority vote -- affects such a compound system's performance. In this paper, we initiate the study of scaling properties of compound inference systems. We analyze, theoretically and empirically, how the number of LM calls affects the performance of \VISFull{} and \FVISFull{}, two of the simplest compound system designs, which aggregate LM responses via majority voting, optionally applying LM filters. We find, surprisingly, that across multiple language tasks, the performance of both  \VISFull{} and \FVISFull{} can first increase but then decrease as a function of the number of LM calls. 
Our theoretical results suggest that this non-monotonicity is due to the diversity of \dataitem{} difficulties within a task: more LM calls lead to higher performance on ``easy'' queries, but lower performance on ``hard'' queries, and non-monotone behavior can emerge when a task contains both types of queries. This insight then allows us to compute, from a small number of samples, the number of LM calls that maximizes system performance, and define an analytical scaling model for both systems. Experiments show that our scaling model can accurately predict the performance of \VISFull{} and \FVISFull{} systems and thus find the optimal number of LM calls to make.
\end{abstract}

\section{Introduction}
Compound AI systems that perform multiple Language Model (LM) calls and aggregate their responses are increasingly leveraged to solve language tasks~\cite{compound-ai-blog,du2023improving,team2023gemini,trinh2024solving,wang2022self}. For example, Google's Gemini Ultra achieved state-of-the-art results on MMLU using a CoT@32 voting strategy: the LM is called 32 times, and then the majority vote of the 32 responses is used in the final response~\cite{team2023gemini}.
Other compound systems filter responses using an LM before selecting one~\cite{AlphaCode2}.
%In addition to their superior performance, compound systems are appealing because they are convenient to construct and debug~\cite{compound-ai-blog}.
 %Invoking multiple LM calls also enhances  compound systems' robustness to failures of individual LM calls.

A natural question is, thus, how does scaling the number of LM calls affect the performance of such compound systems? This question is under-explored in research,  but
characterizing the scaling dynamics is crucial for researchers and practitioners to estimate how many LM calls are needed for their applications and allocate computational resources aptly.
Understanding these scaling dynamics is also helpful in recognizing the limits of compound inference strategies.
%{\color{blue} To better understand such compound systems, it is important to determine how their performance scales with the number of LM calls. This dynamic is surprisingly under-explored, but useful for understanding the capabilities, limitations, and computational costs of deploying these systems in practice. }

\begin{figure}[t]
    \centering
    \includegraphics[width=\textwidth]{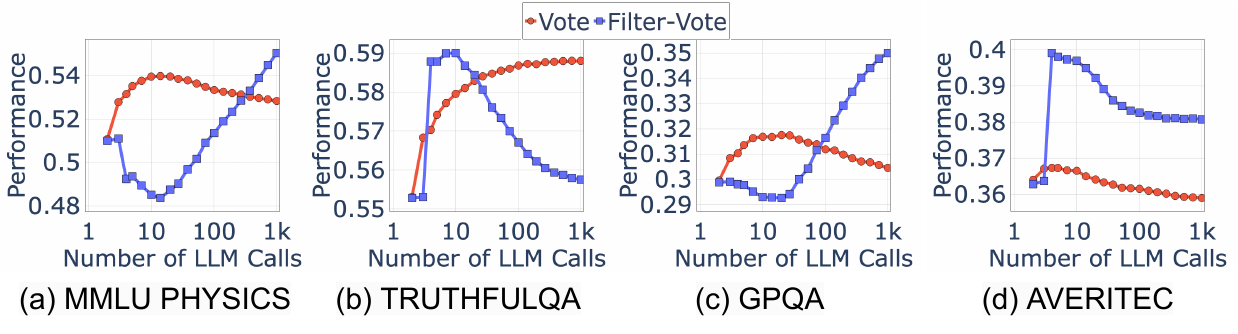}
    \caption{The scaling behavior of \VISFull{} and \FVISFull{}. Interestingly, their performance is often a non-monotonic function of the number of LM calls. For example, as the number LM calls increases, \VISFull{}'s performance initially increases but then decreases, while \FVISFull{}'s performance initially decreases but then increases on the MMLU PHYSICS dataset. }
    \label{fig:ScaleLaw:Behavior}
\end{figure}

 As a first step towards answering this question, 
we study the scaling properties of two popular compound system designs, \VISFull{} and \FVISFull{}. \VISFull{} aggregates multiple proposed answers via majority vote, as in Gemini's CoT@32. \FVISFull{} leverages a filter before performing a majority vote, similar to  AlphaCode 2~\cite{AlphaCode2}.
While the inference system design space is broad, we focus on \VISFull{} and \FVISFull{} for two reasons. First, they have already been used in real applications, such as Gemini's CoT@32 strategy. Second, despite their simplicity, they exhibit nontrivial scaling properties.  
Specifically, although one might expect their performance to monotonically increase as more LM calls are invoked,  we have identified a surprising phenomenon, across multiple language tasks, exhibited by these systems: growing the number of LM calls initially improves performance but then degrades it, as shown in Figure \ref{fig:ScaleLaw:Behavior}. 
% {\color{blue}As a first step towards answering this question, we study the scaling properties of one-layer Voting \LMNets{}, a particularly simple compound system that aggregate multiple proposed answers by majority vote. While \LMNets{} such as AlphaCode 2 uses more sophisticated aggregation mechanisms, we focus on one-layer Voting  \LMNets{} for two reasons. First, they already represent real compound systems such as Gemini evaluation on MMLU. Second, despite their  simplicity, they exhibit non-trivial scaling properties.  

% Specifically, although one might expect the performance of \LMNets{} to monotonically increase as more LM calls are invoked, we find empirically on several language tasks an unexpected phenomenon. Increasing the number of LM calls initially leads to improved performance of \LMNets{}, but further increases degrade performance (see  Figure \ref{fig:ScaleLaw:intro1}). }

\begin{figure}[t]
    \centering
    \includegraphics[width=0.99\textwidth]{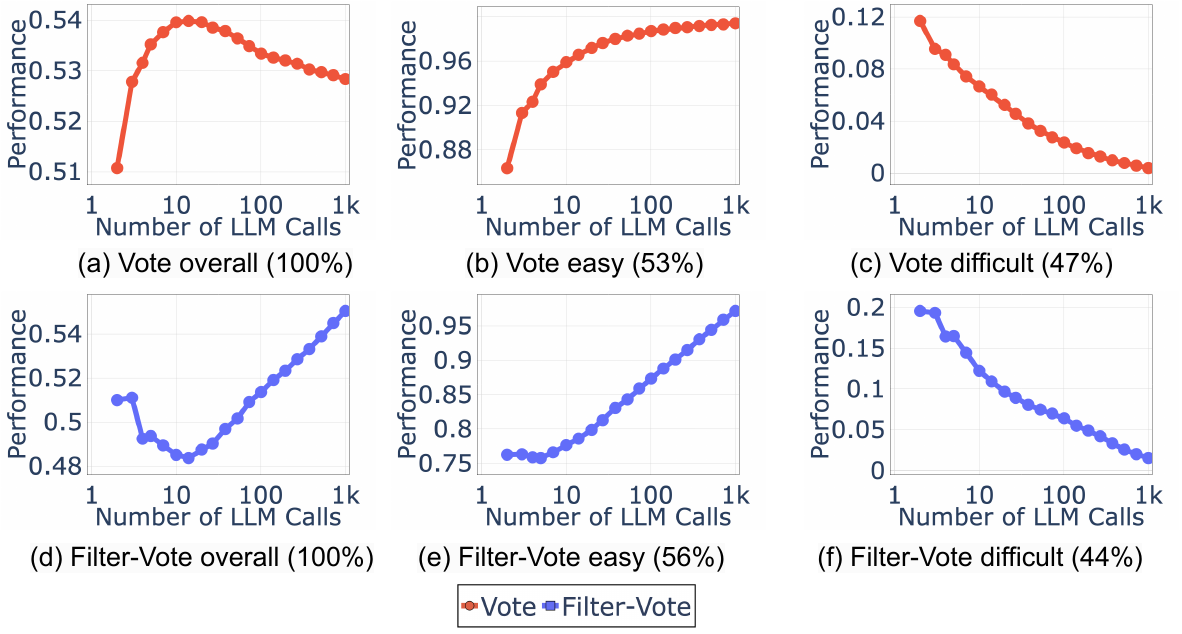}
    \caption{Performance breakdown on MMLU PYHSICS. As the number of LM calls increases, \VISFull{} and \FVISFull{} perform increasingly better on easy queries but increasingly worse on difficult ones. %About 47\% queries in PHYSICS are considered difficult for \VISFull{}, while for \FVISFull{} about 44\% queries are difficult.
    }
    \label{fig:ScaleLaw:intro}
\end{figure}

This surprising effect motivates our theoretical study of \VISFull{} and \FVISFull{}, which explains this non-monotone effect through the diversity of \emph{\dataitem{}  difficulty} in a given task (see Figure \ref{fig:ScaleLaw:intro} and Theorem \ref{thm:scaleLaw:character}). At a high level, our results show that more LM calls continuously lead to better performance on ``easy'' queries and worse performance on ``hard'' queries. When a task is some mixture of  ``easy'' and ``hard'' queries, the non-monotone aggregate behavior emerges.  Formally, a query is easy if a compound system with infinitely many LM calls gives a correct answer and hard otherwise. We provide mathematical conditions under which a query is easy or difficult for \VISFull{} or \FVISFull{}.  We also derive a performance scaling model for both systems that explicitly models \dataitem{} difficulties, and present an algorithm that lets users fit the scaling law's parameters using a small number of samples. In experiments with GPT-3.5, we show that these algorithms can let us estimate the scaling behavior for various problems and identify the optimal number of calls to make to the LM to maximize accuracy. Our main contributions are:

\begin{mybox}

\textbf{Non-monotonic Scaling Behavior.} We find empirically that the performance of \VISFull{} and \FVISFull{} is often non-monotonic in the number of LM calls.\\%We discover a non-monotonic scaling behavior of compound inference systems. Specifically, as the number of LM calls increases, the performance of \VISFull{} and \FVISFull{} can initially increase and then decrease empirically. \\

\textbf{\Dataitem{} Difficulty-based Explanation.} We propose a formal notion of \dataitem{} difficulty. We then argue empirically and show in a simple analytical model that the diversity of \dataitem{} difficulty can explain the scaling behavior of  \VISFull{} and \FVISFull{}. Additional LM calls improve performance on easy queries and degrades performance on difficult queries. We provide precise conditions under which the non-monotone scaling behavior emerges on datasets containing both easy and difficult queries in our analytical model.\\%We propose a formal notion of \dataitem{} difficulty and explain the scaling behavior by the diversity of \dataitem{} difficulty. In particular, we show theoretically and empirically that more LM calls increment the performance on easy queries but downgrade that on difficult queries. We give mathematical conditions under which the non-monotone scaling behavior emerges on datasets containing both easy and difficult queries. \\

\textbf{Heuristic for Optimal Number of LM Calls.} We empirically validate predictions for the optimal number of LM calls from our analytical model for \VISFull{} and \FVISFull{}, which can be estimated from a small number of queries.

%\textbf{Analytical Scaling Model.} We design the first analytical performance model for \VISFull{} and \FVISFull{} inference strategies. %Our model is an integral of the performance on each query, which is negative exponential for easy queries and exponential for hard ones. 
%Experiments show that our model can accurately predict the nonmonotone scaling behavior and allows us to optimize the number of LM calls based on a small number of queries. 
\end{mybox}

%Our study on \VISFull{}  and \FVISFull{} opens the door for understanding and leveraging the scaling properties of compound AI systems. We study here mainly \VISFull{} that use a majority voting scheme and \FVISFull{} that uses LM filters to aggregate LM generations, but many compound systems leverage different aggregation mechanisms designed for different tasks. For example, Medprompt adopts an LM as a judge to select answers for medical question answering~\cite{nori2023capabilities}, and AlphaGeometry combines LM calls with exact derivations from a symbolic engine~\cite{trinh2024solving}. %The scaling laws of these systems will differ from \VISFull{} and remain an interesting open problem. %Another interesting future direction is to estimate item difficulties accurately and vary the number of inference calls for each item type in a voting system.

%This is of particular practical values as integrating an accurate predictor in a compound system can directly improve the overall performance. %For instance, 

In a nutshell, our work shows that more LM calls do not necessarily improve the performance of compound AI systems and that it is possible{\color{black}, at least in some cases,} to predict how the number of LM calls affects AI systems' performance and thus decide the optimal number of LM calls {\color{black} for a given task}. Our work focuses on tasks with a fairly small number of possible responses (e.g., multiple-choice questions) that support a majority vote, but tasks with many valid outputs, such as chat, remain under-explored. We will release the code and datasets used in this paper, and hope to excite more research regarding the scaling properties of compound AI systems.

%TODO: rename the technique by voting inference networks. OK to keep the title. 

%There are many possible ways to use LM calls. List some of the works. One option is majority vote, google gemini; verification is another option, alpha code.
%Other options, there are many types. 

%A discussion on the other types of inference networks. E.g.,a perfect judge-based system.  A natural task with the majority voting network.

%TODO: inference networks to inference systems? We should say there are many versions of inference networks beyond majority voting ones.

%Maybe good to start with Google gemini that uses majority voting. 

%Boris: Study the simplest settings and find non-trivial phenomenon; This motivates us to study more complicate systems in the future. 

%Boris: We found show an empirical result for other network system;  No need to analyze it, but just say that this is not brittle;  maybe only to the majority vote? 

\section{Related Work} 
\paragraph{Neural Scaling Laws.}
There has been extensive research on how the training parameters affect the performance of neural language models~\cite{kaplan2020scaling,sorscher2022beyond,bahri2021explaining,mckenzie2023inverse,isik2024scaling}. Among others, the model loss has been empirically shown to follow a power law as the number of model parameters and training tokens~\cite{kaplan2020scaling}. Researchers have also proposed theories to explain the empirical scaling laws~\cite{bahri2021explaining}. By contrast, recent work has found that scaling up the model parameters leads to performance decay on certain tasks~\cite{mckenzie2023inverse}. To the best of our knowledge, there is no study on how the number of LM calls affects the performance of a compound system, which is complementary to scaling laws on training parameters and data set sizes.

\paragraph{Compound Systems using LMs.} Many inference strategies that perform multiple model calls have been developed to advance performance on various language processing tasks~\cite{xi2023rise,du2023improving,team2023gemini,trinh2024solving,wang2022self,chen2023frugalgpt,zhang2023ecoassistant,vsakota2023fly}. For example, Gemini reaches state-of-the-art performance on MMLU via its CoT@32 majority voting scheme~\cite{team2023gemini}. Self-consistency~\cite{wang2022self} boosts the performance of chain-of-thought via a majority vote scheme over multiple reasoning paths generated by PaLM-540B.
Finally, AlphaCode 2~\cite{AlphaCode2} matches the 85th percentile of humans in a coding contest regime by generating up to one million samples per problem via an LM and then filtering the answer set down. 
While these approaches are empirically compelling, there has been little systematic study of how the number of LM calls affects these systems' performance, and how it should be tuned for a given application.

\paragraph{Compound System   Evaluation.} Compound AI systems are increasingly evaluated in traditional benchmarks~\cite{cobbe2021training,hendrycks2021measuring,thorne2018fever,nie2019adversarial} as well as domain-specific new datasets~\cite{katz2024gpt, mehandru2024evaluating, kung2023performance} and the agent environments~\cite{liu2023agentbench,shridhar2020alfworld,jimenez2024swebench}. We refer the interested readers to a survey for more details~\cite{chang2024survey}. Existing papers focus on obtaining a single metric of a given system, while our goal is to understand how the a compound system's performance is affected by its parameters (e.g., \# of LM calls).

%\paragraph{LM Ensembles.} FrugalGPT~\cite{chen2023frugalgpt,zhang2023ecoassistant}, Blender, 

%\paragraph{Majority Voting.}

\section{Inference System Designs}\label{LMNet:sec:theory}
\begin{figure}[t]
    \centering
    \begin{minipage}{0.48\textwidth}
        \begin{algorithm}[H]
            \caption{\VISFull{}. }\label{alg:scalelaw:infernet}
            \KwIn{A user query $x$, \#  gen responses $K$}
            \KwOut{A response $\hat{y}$}
            Sample $\theta_1, \theta_2,\cdots, \theta_K$ i.i.d. from  $\Theta$\;
            Generate $z_{k} = G(x,\theta_k), k=1,\cdots, K$\;
            
            Set $\hat{y}_K = V(z_1,\cdots, z_K) $\;
            Return $\hat{y}_K$
        \end{algorithm}
    \end{minipage}\hfill
    \begin{minipage}{0.49\textwidth}
        \begin{algorithm}[H]
            \caption{\FVISFull{}. }\label{alg:scalelaw:infernetFVIS}
            \KwIn{A user query $x$, \# gen responses $K$}
            \KwOut{A response $\hat{y}$}
            Sample $\theta_1, \theta_2,\cdots, \theta_K$ i.i.d. from  $\Theta$\;
            Generate $z_{k}, e_{k} = G(x,\theta_k), k=1,\cdots, {K}$\;
            Generate $w_k = \Phi(x,z_{k},e_{k}),k=1,\cdots, K $\;
    \If{$\max_k w_k$ = 0}{
        Set $\hat{y}_K = V(z_1,\cdots, z_{K}) $\;
    }
    \Else{
    Set $\hat{y}_K = V(\{\!| z_k \mid w_k = 1 |\!\}
)$\;
    }

            Return $\hat{y}_K$
        \end{algorithm}
    \end{minipage}
\end{figure}
In this paper, we focus on two simple and natural inference system designs: \VISFull{} and \FVISFull{}. \VISFull{} is inspired by and resembles several real-world compound AI systems, such as self-consistency~\cite{wang2022self}, Medprompt~\cite{nori2023capabilities}, and Gemini CoT@32 strategy~\cite{team2023gemini}, while \FVISFull{} represent many other real-world compound AI systems including AlphaCode 2~\cite{AlphaCode2} and AlphaGeometry~\cite{trinh2024solving}. 
Note that this paper focuses on tasks with a small number of possible answers.

\paragraph{Building Blocks.} \VISFull{} and \FVISFull{} rely on three building blocks, a generator $G(\cdot, \cdot)$, a majority voter $V(\cdot)$, and a filter $\Phi(\cdot,\cdot,\cdot)$. The generator $G(\cdot, \cdot)$ takes a user query $x$ and $\theta \in \Theta$ as inputs and produces a candidate answer and an explanation. Here, instantiations of $\Theta$ are a design choice of users and can encode many generation strategies.  
For example, even with a single fixed LM, diverse generations may be achieved by using a non-zero temperature and different prompt wordings or few-shot examples for each call to the LM. If $\Theta$ contains different LMs, then this system definition can also represent LM ensembles.
The majority voter $V$ returns the mode of its input, i.e., $V(z_1,z_2,\cdots, z_K) \triangleq \arg\max_{a\in A} \sum_{k=1}^{K}\mathbbm{1}_{z_k=a}$, and breaks ties arbitrarily. Here, $A$ is the space of all possible answers. Finally, the filter $\Phi(\cdot,\cdot)$ takes the user query and multiple candidate answers as input, and only returns the subset that an LM believes is correct. 

\paragraph{\VISFull{}.}  Given a user query $x$, \VISFull{} (i) first creates $K$ candidate answers by calling the generator $G$, and then (ii) uses the majority voter $V$ to choose one as the final response $\hat{y}_K$. The details are given in Algorithm \ref{alg:scalelaw:infernet}.

\paragraph{\FVISFull{}.} Given a user query, \FVISFull{} (i) first generates multiple candidate answers, (ii) removes a few candidate answers by the filter $\Phi$, and (iii) then uses the majority voter $V$ to choose one from the remaining answers as the final response. If all answers are removed by the filter, then $V$ is applied on the original candidate answers. Algorithm \ref{alg:scalelaw:infernetFVIS} gives the formal description. %For consistency, \FVISFull{} also utilizes $K$ LM calls in total by (i) generating $K/2$ candidate answers and (ii) using $K/2$ LM calls for the filter.

\begin{table}[t]
  \centering
  %\small
  \caption{Notations.}
    \begin{tabular}{|c||c|}
    \hline
    Symbol &
      Meaning
      \bigstrut\\
    \hline
    $x$ &
      an input query
      \bigstrut\\
    \hline
    $y$  &
      the correct answer
      \bigstrut\\
    \hline
    $K$ &
       the number of LM calls \bigstrut\\
    \hline
    $z_k$ &
      the output by one LM call
      \bigstrut\\
    \hline
    $\hat{y}_K$ &
      the output by an inference system using $K$ LM calls
      \bigstrut\\
    \hline
    $D/D_{Tr}$ &
      test dataset/train dataset
      \bigstrut\\
    \hline
    $A$ &
      answer space
      \bigstrut\\
    \hline
    $\alpha$ &
      fraction of easy queries
      \bigstrut\\
    \hline
    $p_1$ &
      probability of $z_k$ being correct for easy queries \eat{\ion{maybe call this $p_e$ and $p_h$ instead of $p_2$ }}
      \bigstrut\\
    \hline
    $p_2$ &
      probability of $z_k$ being correct for difficult queries
      \bigstrut\\
    \hline
    $F(K;D)$ &
      Accuracy of an \LMNet{} with $K$ LM calls per query  on $D$ \bigstrut\\
    \hline
    $G(K;D)$ &
     Analytical Performance Model (to approximate $F(K;D) $) 
      \bigstrut\\
    \hline
    \end{tabular}%
  \label{tab:ScaleLaw:Notations}%
\end{table}%

\section{Analytical Model of Scaling Behavior}
Now we present our analytical performance model of \VISFull{} and  \FVISFull{} strategies. Specifically, we are interested in understanding the behavior of $F(K;D)\triangleq \mathbbm{E}[\hat{y}_K=y]$, where the expectation is over $D$ and the candidate responses. All notations are summarized in Table \ref{tab:ScaleLaw:Notations}.

\subsection{When do more LM calls lead to an increase or decrease in performance?}
\begin{figure}[t]
    \centering
    \includegraphics[width=0.99\textwidth]{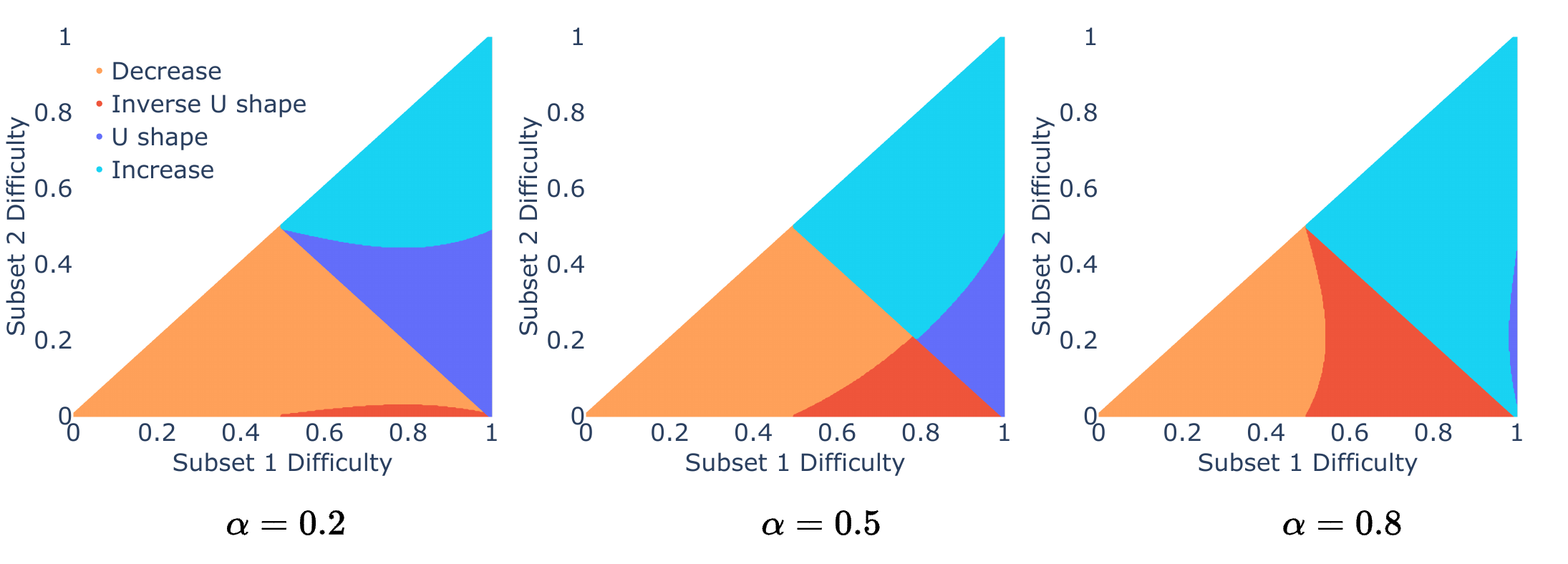}
    \caption{\eat{\ion{Increase the patches of color in the legend. It is hard to see which colors are assigned to Decrease/Inverse U shape/...}} How the \dataitem{} difficulties shape the landscape of a one-layer Voting \LMNet{}'s performance. Informally, if the overall task is ``easy'' ($p_1+p_2>1$), but the fraction of ``hard'' queries is large ($\alpha<1-\frac{1}{t}$), then as the number of LM calls increases, the Voting \LMNets{}' performance increases first but then decreases. We call such a landscape a ``inverse U shape''. 
Similarly,  if the overall task is ``hard'' ($p_1+p_2<1$), but the fraction of ``hard'' queries is small ($\alpha>1-\frac{1}{t}$), then enlarging the number of LM calls leads an initial decrease and then increase. Such a landscape is called a ``U shape''. When $\alpha$ is large, the U-shape is less likely to occur while the inverse U-shape becomes more common. Smaller $\alpha$ leads to an opposite trend. %\ion{Why not label the plots in Figure 2 using the shape names you use here, i.e., Increasing, Decreasing, Inverse U Shape, etc? Will make this figure a little bit easieer to follow.}
}
    \label{fig:ScaleLaw:Landscape}
\end{figure}
Our first key insight is that individual query difficulty is crucial in LM calls' effects. To see this, let us first introduce \dataitem{} difficulty indicator. %\ion{Why use both "query" and "\dataitem{}". Why not only one?}
\begin{definition}
Given a user query $x$, $d(x)$ is called an \dataitem{} difficulty indicator if 
    \begin{equation*}
\lim_{K\rightarrow\infty} F(K,x)=\begin{cases}
    0 & \textit{iff }  d(x)>0, \\
    1 & \textit{iff }  d(x)<0
\end{cases}
\end{equation*}
\end{definition}
Intuitively, a positive \dataitem{} difficulty indicator implies the query is difficult, i.e., infinitely many LM calls lead to an incorrect final answer, and a negative value implies the query is easy, i.e., infinitely many LM calls eventually give a correct final answer. For simplicity, we assume that the limit of $F(K,x)$ is always either 0 or 1 for \VISFull{} and \FVISFull{}, i.e., eventually the answer is correct or incorrect. Also note that $d(x)$ is scale-invariant, i.e., if $d(x)$ is an item difficulty indicator, then for any positive scalar $\gamma>0$, $\gamma\times d(x)$ is also a difficulty indicator. We will call a query $x$ difficult (easy) if $d(x)>0$ ($d(x)<0$). We give two concrete instantiations as follows.

\begin{lemma}\label{lemma:scalelaw:difficultcondition}
For \VISFull{}, $d_V(x)\triangleq \max_{a\not=y} \Pr[G(x,\theta)=a] - \Pr[G(x,\theta)=y]$ is an \dataitem{} difficulty indicator. For \FVISFull{}, denote $G(x,\theta)=[G_1(x,\theta), G_2(x,\theta)]$. Then $d_F(x)\triangleq \max_{a\not=y} \Pr[G_1(x,\theta)=a| \Phi(x,G_1(x,\theta),G_2(x,\theta))=1]-\Pr[G_1(x,\theta)=y| \Phi(x,G_1(x,\theta),G_2(x,\theta))=1]$ is a \dataitem{} difficulty indicator.
\end{lemma}
Here, $d_V(x)>0$ indicates that an incorrect answer is more likely to be generated than the correct one, hence the query is difficult. Similarly, $d_F(x)>0$ implies that an incorrect answer is more likely to be kept in the filtered answer set.

More LM calls elicit higher performance on easy queries, but lower performance on difficult queries. Therefore, the performance, $F(K;D)$, is more difficult to characterize when the data set $D$ contains both easy and difficult queries. Here, we study \VISFull{} on a special case of $D$ to understand how the difficulty impacts the performance function $F(K;D)$.   

\paragraph{A case study on a specific dataset.} Let us consider a specific dataset $D_{\alpha,p_1,p_2}$ with answer space cardinality $|A|=2$. Here, $\alpha\in[0,1]$ queries in $D$ are   $x_1$ such that $Pr[G(x_1,\theta)=y]=p_1>\frac{1}{2}$, and $1-\alpha$ queries are $x_2$ such that $Pr[G(x_2,\theta)=y]=p_2<\frac{1}{2}$. The following theorem qualitatively characterizes the performance of \VISFull{} on this dataset. 

\begin{theorem}\label{thm:scaleLaw:character}
Let $t\triangleq \frac{p_2 (1-p_2) (\frac{1}{2}-p_2)}{p_1 (1-p_1) (p_1-\frac{1}{2})}+1$. If  $p_1+p_2\not=1$ and $K$ is odd, then $F(K;D_{\alpha,p_1,p_2})$ 
   \begin{itemize}
       \item  increases monotonically, if $p_1+p_2>1$ and  $\alpha \geq 1-\frac{1}{t}$
       \item decreases monotonically,  if $p_1+p_2<1$ or $\alpha\leq 1-\frac{1}{t}$
       \item increases and then decreases, if $p_1+p_2>1$ and $\alpha<1-\frac{1}{t}$
       \item decreases and then increases, if $p_1+p_2<1$ and $\alpha>1-\frac{1}{t}$
   \end{itemize}
   \end{theorem}

Theorem \ref{thm:scaleLaw:character} precisely connects the \dataitem{} difficulty with the performance landscape. Here, $t$ is a constant that only depends on $p_1$ and $p_2$, i.e., the probability of an LM's generation being correct on easy and hard queries, respectively. Intuitively, $t$ quantifies the difficulty similarity between the easy and hard queries: it becomes larger if the easy queries are more difficult ($p_1$ is smaller) or the hard queries are less difficult ($p_2$ is larger).  Interestingly, it suggests that, for some \dataitem{} difficulty distribution, a non-monotone effect of the number of LM calls is expected. Informally, if the overall task is ``easy'' ($p_1+p_2>1$), but the fraction of ``hard'' queries is large ($\alpha<1-\frac{1}{t}$), then as the number of LM calls increases, the Voting \LMNets{}' performance increases first but then decreases. We call such a landscape a ``inverse U shape''. 
Similarly,  if the overall task is ``hard'' ($p_1+p_2<1$), but the fraction of ``hard'' queries is small ($\alpha>1-\frac{1}{t}$), then enlarging the number of LM calls leads an initial decrease and then increase. Such a landscape is called a ``U shape''. 
This well explains the U-shape of \LMNets{}' performance shown in Figure \ref{fig:ScaleLaw:Behavior}.Figure \ref{fig:ScaleLaw:Landscape} visualizes the effects of \dataitem{} difficulty on the performance landscape in more detail. 

\subsection{What is the analytical scaling model?}
Now we derive an analytical scaling model for both \VISFull{} and \FVISFull{}. Noting that the performance is the average of $F(K,x)$ for each $x$ in the dataset, the key challenge is identifying the shape of $F(K,x)$ for easy and difficult queries. Let us first consider the special case  $|A|=2$, where we can obtain a close form result for \VISFull{}.

\begin{theorem}\label{thm:scalelaw:scalingmodel}
If $|A|=2$, then on any query $x$, the performance of \VISFull{} is $F(K,x) = I_{\frac{1-d_V(x)}{2}}(\frac{K+1}{2},\frac{K+1}{2})$, where $I_x(a,b)\triangleq \int_{0}^{x} t^{a-1} (1-t)^{b-1}dt/\int_{0}^{1} t^{a-1} (1-t)^{b-1}dt$ is the regularized incomplete beta function.
\end{theorem}
Thus, for \VISFull{} with $|A|=2$, $F(K,D)=\mathbbm{E}_{x\sim D}[I_{\frac{1-d_V(x)}{2}}(\frac{K+1}{2},\frac{K+1}{2})]$.

How about \FVISFull{} and the general answer space? Admittedly, an exact scaling model is challenging to obtain. Instead, we give an approximation model inspired by the special case. We first note that $F(K,x)$ should be treated separately for difficult and easy queries: after all, as a function of $a$, the incomplete beta function $I_x(a,a)$ monotonically increases/decreases if $x>\frac{1}{2}$ ($x<\frac{1}{2}$). Second, $I_x(a,a)$ grows roughly exponentially in $x$, and this trend should hold for general answer space and for both \VISFull{} and \FVISFull{}. Hence, we propose the following  scaling model
\begin{equation*}
    G(K,x) \triangleq \begin{cases}
        e^{-c_1(x) K - c_2(x) \sqrt{K}+c_3(x)}, & \textit{if } d(x)>0, \\
        1-e^{-c_1(x) K - c_2(x) \sqrt{K}+c_3(x)},  & \textit{if } d(x)<0
    \end{cases}
\end{equation*}
where constants $c_1(x)>0, c_2(x)>0, c_3(x)$ do not depend on the number of LM calls $K$. Therefore, our analytical performance scaling model is $G(K,D) = \mathbbm{E}_{x\sim D}[G(K,x)]$. In practice, one can use a training dataset $D_{Tr}$ to fit the parameters in $G(K,D)$. Note that given a query $x$, the parameters $c_i(x)$ can be different for \VISFull{} and \FVISFull{}. In particular, if the filter is of high quality, then the performance should converge quickly, and thus the constants $c_i(\cdot)$ are likely to be larger. Otherwise, the performance should scale slower, and thus the constants $c_i(\cdot)$ should be smaller.  We will show in the experiments that $G(K,D)$ matches the empirical performance $F(K,D)$ accurately.

\subsection{How to optimize the number of LM calls?}
In general, one can always (i) fit the analytical scaling model $G(K,D)$, and (ii) then use $\max_K G(K,D)$ to obtain the optimal number of LM calls. Interestingly, we show that for a special case, we can derive the optimal number of LM calls. %\ion{... we can derive the optimal number of LM calls.}
\begin{theorem}\label{thm:scalelaw:optimalnetworksize}
If $p_1+p_2>1$ and $\alpha<1-\frac{1}{t}$, then the number of LM calls  $K^*$ that maximizes $F(K,D_{\alpha,p_1,p_2})$ for \VISFull{} 
 (up to rounding) is 
\begin{align*}
    K^* = 2 \frac{\log\frac{\alpha}{1-\alpha}\frac{2p_1-1}{1-2p_2}}{\log \frac{p_2(1-p_2)}{p_1(1-p_1)}} 
\end{align*}
\end{theorem}
The optimal number of LM calls depends on the \dataitem{} difficulty. For example, $K^*$ will be larger if $\alpha$ grows (up to $1-\frac{1}{t}$). 
That is, if there are more ``easy'' queries than ``difficult'' queries, then more LM calls should be adopted. %This suggests that a difficulty-aware design of Voting \LMNets{} may offer better performance than a difficulty-agnostic design. We will analyze $K^*$ in detail in Section \ref{sec:scalelaw:exp_optimalsize}. 

\section{Experiments}\label{sec:scalelaw:exp}
We compare the empirical performance of \VISFull{} and \FVISFull{} and the performance predicted by our analytical scaling model.  Our goal is three-fold: (i) validate that there are cases where more LM calls do not monotonically improve the performance of these \LMNets{}, (ii) justify that the number of LM calls has opposite effects on easy and difficult queries, and (iii) explore whether our analytical scaling model can accurately predict the performance of \VISFull{} and \FVISFull{}, and thus guide the design of \LMNets{} such as optimizing number of LM calls.

\paragraph{Datasets and LM.} To understand the scaling properties of \VISFull{} and \FVISFull{}, we conduct systematical experiments on both (i) real-world datasets and (ii) synthetic datasets with controlled \dataitem{} difficulties.  Specifically, the real-world datasets include  MMLU PHYSICS~\cite{hendrycks2020measuring}, TRUTHFULQA~\cite{lin2021truthfulqa}, GPQA~\cite{rein2023gpqa}, and  AVERITEC~\cite{schlichtkrull2024averitec}. MMLU PHYSICS contains high school physics questions extracted from the original MMLU dataset. TRUTHFULQA measures whether a language model is truthful in generating answers to questions. GPQA queries are generated by experts in biology, physics, and chemistry. Each query in AVERITEC is a claim and the goal is to verify its correctness. Each query is prompted as a multiple-choice question for objective evaluation. The details of these datasets and prompts can be found in the Appendix. The synthetic dataset is $D_{\alpha,p_1,p_2}$ as introduced in Section \ref{LMNet:sec:theory}, and we study the scaling behavior by varying the parameters. We use GPT-3.5-turbo-0125 on the real-world datasets.  All experiments are averaged over 1,000 runs.

\paragraph{Non-monotonic Scaling Behavior.} We start by understanding how the number of LM calls affects the performance of \VISFull{} and \FVISFull{} empirically. As shown in Figure \ref{fig:ScaleLaw:Behavior}, we observe a non-monotonic behavior: more LM calls can sometimes lead to a drop in performance! This underscores the importance of scaling performance modeling.

\paragraph{A case study on AVERITEC.} Now let us perform a case study on the AVERITEC dataset to understand the intriguing behavior better. In particular, we use the deployment partition of  AVERITEC~\cite{schlichtkrull2024averitec}, which contains 500 fact verification questions. The goal is to determine if a given claim should be (A) refused, (B) supported, or there is (C) conflicting evidence or (D) not enough evidence. We evaluate the performance of both \VISFull{} and \FVISFull{} on AVERITEC. In addition, we fit our analytical scaling model with 2, 5, 10, 20, 50, 100 LM calls. Then we use it to predict the performance of \VISFull{} and \FVISFull{} using 100 randomly drawn number of LM calls.
\begin{figure}[t]
    \centering
    \includegraphics[width=0.99\textwidth]{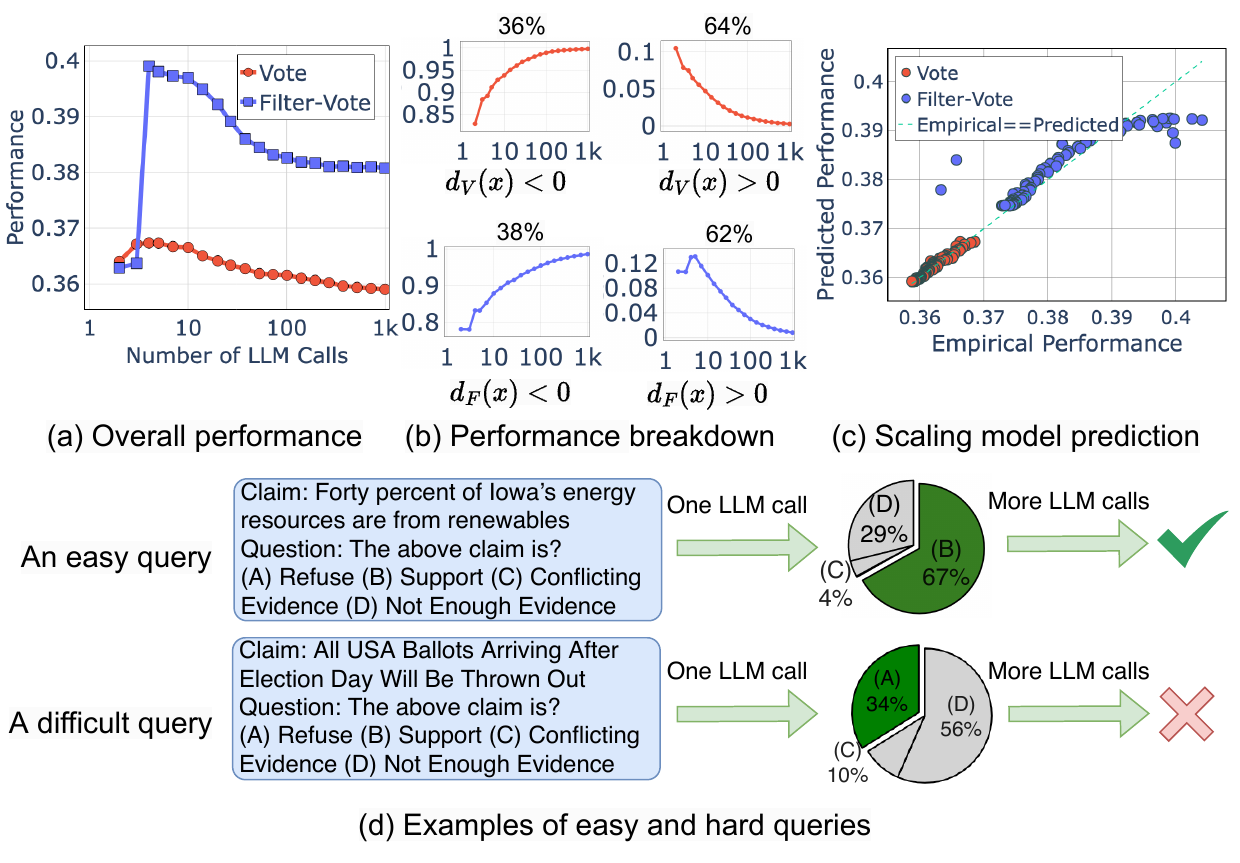}
    \caption{A case study on the AVERITEC dataset. (a) As more LM calls are invoked, the overall performance of \VISFull{} and \FVISFull{} both initially increases but then decreases. (b) This U-shape can be perfected explained by the opposite effects on easy and difficult queries: More LM calls lead to higher performance on easy queries, but lower performance on difficult ones. (c) Our analytical scaling model accurately predicts the empirical performance. (d) Examples of an easy query and a difficult one. One LM call gives the correct answer with probability higher than any other answers (67\%), and thus \VISFull{} with more calls eventually gives the correct answer. For the difficult query,  the probability of the correct answer (34\%) is lower than that of an incorrect answer (56\%). Thus, \VISFull{} with more LM calls eventually always generates a wrong answer.}
    \label{fig:LMNet:casestudy}
\end{figure}

As shown in Figure \ref{fig:LMNet:casestudy}, there are several intriguing behaviors. First, more LM calls do not always lead to better performance. In fact, the performance of  both \VISFull{} and \FVISFull{} increases first but then decreases as the number of LM calls increases from 2 to 1000 (Figure \ref{fig:LMNet:casestudy}(a)). The performance breakdown (shown in Figure \ref{fig:LMNet:casestudy}(b)) gives a natural explanation. More LM calls lead to higher performance when a query is relatively difficult ($d_F(x)>0$ for \FVISFull{} and $d_V(x)>0$ for \VISFull{}), but lower performance when a query is relatively easy ($d_F(x)<0$ for \FVISFull{} and $d_V(x)<0$ for \VISFull{}). 
We also observe a high correlation between the performance predicted by our proposed analytical scaling model and the empirical performance, as depicted in Figure \ref{fig:LMNet:casestudy} (c). This implies the optimal number of LM calls can also be accurately predicted by our scaling model.
Finally, we also give examples of one easy query and one difficult query in  Figures \ref{fig:LMNet:casestudy} (d). On the easy example, one LM call gives the correct answer with a probability higher than any other answer (67\%). Thus, more LM calls eventually give the correct answer. On the difficult query,  the probability of the correct answer (34\%) is lower than that of an incorrect answer (56\%). Thus, \VISFull{} with more LM calls eventually always generates a wrong answer.

\begin{figure}[t]
    \centering
    \vspace{0mm}
\includegraphics[width=0.99\textwidth]{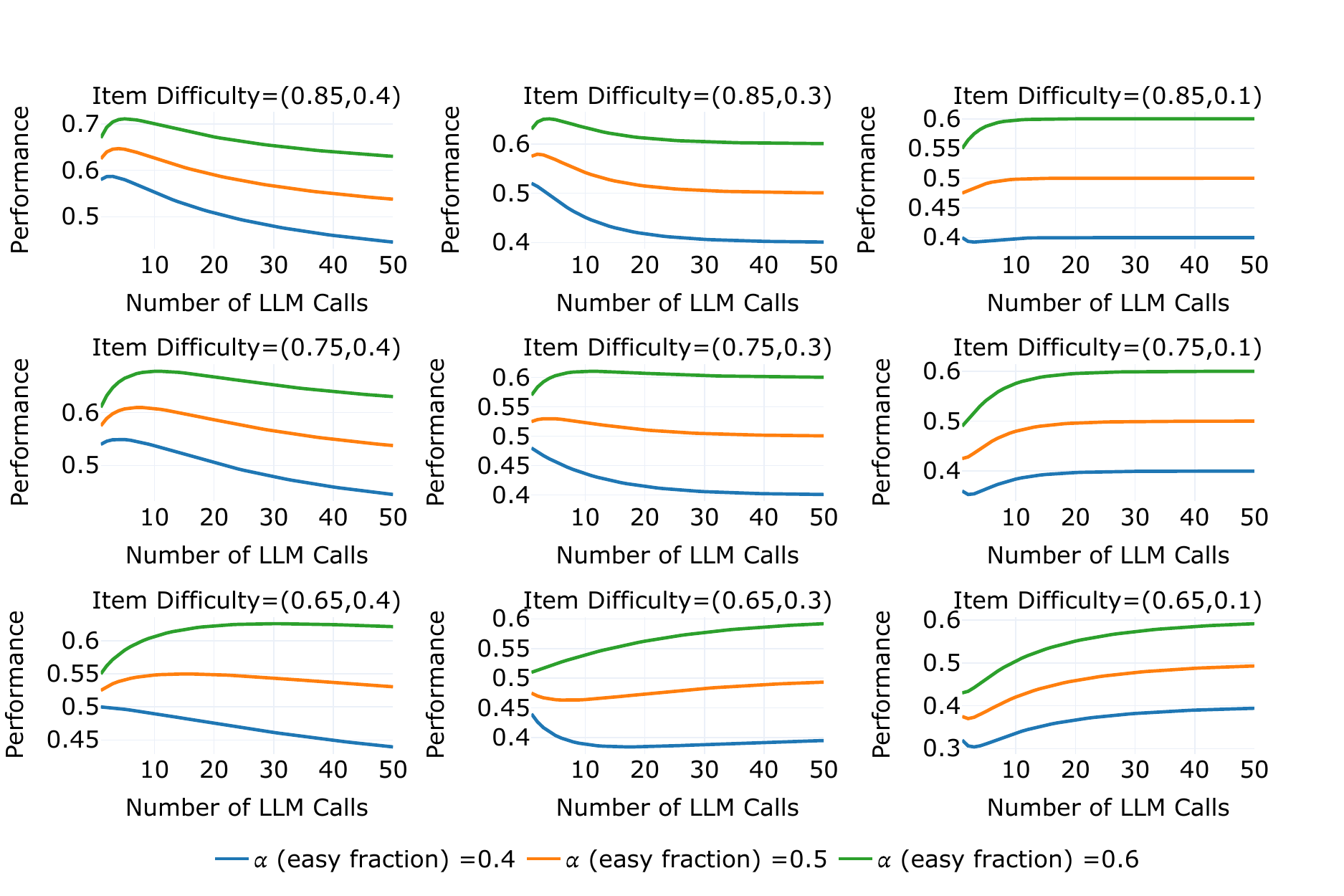}
    \caption{Overall performance of one-layer Voting \LMNets{} on synthetic datasets with bi-level difficulty. Overall, we observe that increasing the number of LM calls does not necessarily lead to performance improvements. For example, when the \dataitem{} difficulty is $(0.85, 0.4)$, the performance increases first and then decreases as the number of LM calls grows. When the \dataitem{} difficulty is $(0.65,0.1)$, there is a reverse trend: the performance goes down first and then goes up. This validates our empirical observation that a larger number of LM calls and thus more resources may not necessarily result in better performance.}
    \label{fig:ScaleLaw:Simulation}
\end{figure}

\paragraph{Difficulty distribution determines whether more LM calls help.}
To systematically understand how the \dataitem{} difficulty affects \LMNets{}' performance landscape, we synthesize a bi-level difficult dataset $D_{\alpha,p_1,p_2}$, vary  (i) the fraction of the easy subset $\alpha$ and (ii) the query difficulty $p_1, p_2$, and then study the scaling performance of \VISFull{}. When it is clear from the context, we may also call $(p_1, p_2)$ \dataitem{} difficulty.

As shown in Figure \ref{fig:ScaleLaw:Simulation},  we observe that \dataitem{} difficulty plays an important role in the number of LM calls' effects. For example, when the difficulty parameter is $(0.85, 0.1)$ and the fraction of easy queries is $\alpha=0.6$, \VISFull{}'s performance is monotonically increasing as the number of LM calls grows. However, adding more hard queries by changing the fraction $\alpha=0.4$ changes the trend: the performance goes down first for small call numbers and then goes up for larger numbers of calls. It is also interesting to notice an inverse ``U''-shape. For example, when \dataitem{} difficulty is $(0.85,0.4)$, there is a clear U-shape performance. Overall, this justifies that (i) there are cases where more LM calls is not beneficial, and (ii) the diversity of \dataitem{} difficulty critically determines these cases.

\paragraph{Analytical scaling model predicts the optimal number of LM calls.} 
Identifying the optimal number of calls is an important implication of the scaling properties. Here, we compare the optimal number of calls for \VISFull{} predicted by our analytical model and the optimal numbers empirically observed on the bi-level difficult dataset  $D_{\alpha,p_1,p_2}$.  
As summarized in 
Table \ref{tab:ScaleLaw:OptimalNetworkSize} (in the Appendix), the predicted optimal number is exactly the observed optimal number of LM calls, for all \dataitem{} difficulties evaluated. This validates the assumptions made by Theorem \ref{thm:scalelaw:optimalnetworksize}.

\section{Conclusion}
In this paper, we systematically study how the number of LM calls affects the performance of two natural inference strategy designs: majority voting (\VISFull) and majority voting after filtering results with an LM (\FVISFull). We find that increasing the number of LM calls can lead to non-monotone behavior, \emph{e.g.,} first increasing performance and then decreasing it. We offer theoretical analysis that attributes this phenomenon to the diversity of query difficulties within a task, and conduct experiments to validate our analysis.
Furthermore, we show how to estimate the optimal number of LM calls to make for a given task using a small number of queries to fit the parameters of our analytical model, thus helping practitioners optimize their system designs.
Note that in this work, we do not discuss the cost of LM calls; this is an important dimension to weigh in practice, and in future work we will investigate how to balance cost, performance, and latency scaling. Overall, our study shows that more LM calls are not necessarily better and underscores the importance of compound system design. We hope our findings and analysis will inspire more research into maximally effective AI system construction.

% \james{Good to also explain that in this work we don't consider the cost of LM calls; that's clearly important in practice and in future work we will investigate how to balance cost and performance scaling.}

%Our findings open the door to understanding how compound systems scale with the number of LM calls, and pinpoint many interesting future directions to study. 
%This study focuses on one-layer Voting \LMNets{}, but there are many other types of inference systems and, more generally, compound systems. How other constructions scale remains an interesting and open question. For example, rather than a voting-based aggregation of LM responses, a system could rank and select among responses. As a function of the quality of the ranking scheme, very different scaling dynamics can emerge -- for example, a system based upon a perfect ranker that always selects the best response (\emph{e.g.,} by checking whether a generated mathematical proof is correct) would exhibit monotone scaling with the number of generations. 
%Going forward, it is also of high practical value to further study how to accurately predict the difficulty of a given query.

\newpage
\bibliography{sample}

\newcommand{\etalchar}[1]{$^{#1}$}
\begin{thebibliography}{CWW{\etalchar{+}}24}

\bibitem[Alp23]{AlphaCode2}
AlphaCode.
\newblock Alphacode 2 technical report.
\newblock {\em \url{https://storage.googleapis.com/deepmind-media/AlphaCode2/AlphaCode2_Tech_Report.pdf}}, 2023.

\bibitem[BDK{\etalchar{+}}21]{bahri2021explaining}
Yasaman Bahri, Ethan Dyer, Jared Kaplan, Jaehoon Lee, and Utkarsh Sharma.
\newblock Explaining neural scaling laws.
\newblock {\em arXiv preprint arXiv:2102.06701}, 2021.

\bibitem[CKB{\etalchar{+}}21]{cobbe2021training}
Karl Cobbe, Vineet Kosaraju, Mohammad Bavarian, Mark Chen, Heewoo Jun, Lukasz Kaiser, Matthias Plappert, Jerry Tworek, Jacob Hilton, Reiichiro Nakano, et~al.
\newblock Training verifiers to solve math word problems.
\newblock {\em arXiv preprint arXiv:2110.14168}, 2021.

\bibitem[CWW{\etalchar{+}}24]{chang2024survey}
Yupeng Chang, Xu~Wang, Jindong Wang, Yuan Wu, Linyi Yang, Kaijie Zhu, Hao Chen, Xiaoyuan Yi, Cunxiang Wang, Yidong Wang, et~al.
\newblock A survey on evaluation of large language models.
\newblock {\em ACM Transactions on Intelligent Systems and Technology}, 15(3):1--45, 2024.

\bibitem[CZZ23]{chen2023frugalgpt}
Lingjiao Chen, Matei Zaharia, and James Zou.
\newblock Frugalgpt: How to use large language models while reducing cost and improving performance.
\newblock {\em arXiv preprint arXiv:2305.05176}, 2023.

\bibitem[DLT{\etalchar{+}}23]{du2023improving}
Yilun Du, Shuang Li, Antonio Torralba, Joshua~B Tenenbaum, and Igor Mordatch.
\newblock Improving factuality and reasoning in language models through multiagent debate.
\newblock {\em arXiv preprint arXiv:2305.14325}, 2023.

\bibitem[HBB{\etalchar{+}}20]{hendrycks2020measuring}
Dan Hendrycks, Collin Burns, Steven Basart, Andy Zou, Mantas Mazeika, Dawn Song, and Jacob Steinhardt.
\newblock Measuring massive multitask language understanding.
\newblock {\em arXiv preprint arXiv:2009.03300}, 2020.

\bibitem[HBK{\etalchar{+}}21]{hendrycks2021measuring}
Dan Hendrycks, Collin Burns, Saurav Kadavath, Akul Arora, Steven Basart, Eric Tang, Dawn Song, and Jacob Steinhardt.
\newblock Measuring mathematical problem solving with the math dataset.
\newblock {\em arXiv preprint arXiv:2103.03874}, 2021.

\bibitem[IPH{\etalchar{+}}24]{isik2024scaling}
Berivan Isik, Natalia Ponomareva, Hussein Hazimeh, Dimitris Paparas, Sergei Vassilvitskii, and Sanmi Koyejo.
\newblock Scaling laws for downstream task performance of large language models.
\newblock {\em arXiv preprint arXiv:2402.04177}, 2024.

\bibitem[JYW{\etalchar{+}}24]{jimenez2024swebench}
Carlos~E Jimenez, John Yang, Alexander Wettig, Shunyu Yao, Kexin Pei, Ofir Press, and Karthik~R Narasimhan.
\newblock {SWE}-bench: Can language models resolve real-world github issues?
\newblock In {\em The Twelfth International Conference on Learning Representations}, 2024.

\bibitem[KBGA24]{katz2024gpt}
Daniel~Martin Katz, Michael~James Bommarito, Shang Gao, and Pablo Arredondo.
\newblock Gpt-4 passes the bar exam.
\newblock {\em Philosophical Transactions of the Royal Society A}, 382(2270):20230254, 2024.

\bibitem[KCM{\etalchar{+}}23]{kung2023performance}
Tiffany~H Kung, Morgan Cheatham, Arielle Medenilla, Czarina Sillos, Lorie De~Leon, Camille Elepa{\~n}o, Maria Madriaga, Rimel Aggabao, Giezel Diaz-Candido, James Maningo, et~al.
\newblock Performance of chatgpt on usmle: potential for ai-assisted medical education using large language models.
\newblock {\em PLoS digital health}, 2(2):e0000198, 2023.

\bibitem[KMH{\etalchar{+}}20]{kaplan2020scaling}
Jared Kaplan, Sam McCandlish, Tom Henighan, Tom~B Brown, Benjamin Chess, Rewon Child, Scott Gray, Alec Radford, Jeffrey Wu, and Dario Amodei.
\newblock Scaling laws for neural language models.
\newblock {\em arXiv preprint arXiv:2001.08361}, 2020.

\bibitem[LHE21]{lin2021truthfulqa}
Stephanie Lin, Jacob Hilton, and Owain Evans.
\newblock Truthfulqa: Measuring how models mimic human falsehoods.
\newblock {\em arXiv preprint arXiv:2109.07958}, 2021.

\bibitem[LYZ{\etalchar{+}}23]{liu2023agentbench}
Xiao Liu, Hao Yu, Hanchen Zhang, Yifan Xu, Xuanyu Lei, Hanyu Lai, Yu~Gu, Hangliang Ding, Kaiwen Men, Kejuan Yang, et~al.
\newblock Agentbench: Evaluating llms as agents.
\newblock {\em arXiv preprint arXiv:2308.03688}, 2023.

\bibitem[MLP{\etalchar{+}}23]{mckenzie2023inverse}
Ian~R McKenzie, Alexander Lyzhov, Michael Pieler, Alicia Parrish, Aaron Mueller, Ameya Prabhu, Euan McLean, Aaron Kirtland, Alexis Ross, Alisa Liu, et~al.
\newblock Inverse scaling: When bigger isn't better.
\newblock {\em arXiv preprint arXiv:2306.09479}, 2023.

\bibitem[MMA{\etalchar{+}}24]{mehandru2024evaluating}
Nikita Mehandru, Brenda~Y Miao, Eduardo~Rodriguez Almaraz, Madhumita Sushil, Atul~J Butte, and Ahmed Alaa.
\newblock Evaluating large language models as agents in the clinic.
\newblock {\em NPJ digital medicine}, 7(1):84, 2024.

\bibitem[NKM{\etalchar{+}}23]{nori2023capabilities}
Harsha Nori, Nicholas King, Scott~Mayer McKinney, Dean Carignan, and Eric Horvitz.
\newblock Capabilities of gpt-4 on medical challenge problems.
\newblock {\em arXiv preprint arXiv:2303.13375}, 2023.

\bibitem[NWD{\etalchar{+}}19]{nie2019adversarial}
Yixin Nie, Adina Williams, Emily Dinan, Mohit Bansal, Jason Weston, and Douwe Kiela.
\newblock Adversarial nli: A new benchmark for natural language understanding.
\newblock {\em arXiv preprint arXiv:1910.14599}, 2019.

\bibitem[RHS{\etalchar{+}}23]{rein2023gpqa}
David Rein, Betty~Li Hou, Asa~Cooper Stickland, Jackson Petty, Richard~Yuanzhe Pang, Julien Dirani, Julian Michael, and Samuel~R Bowman.
\newblock Gpqa: A graduate-level google-proof q\&a benchmark.
\newblock {\em arXiv preprint arXiv:2311.12022}, 2023.

\bibitem[SGS{\etalchar{+}}22]{sorscher2022beyond}
Ben Sorscher, Robert Geirhos, Shashank Shekhar, Surya Ganguli, and Ari Morcos.
\newblock Beyond neural scaling laws: beating power law scaling via data pruning.
\newblock {\em Advances in Neural Information Processing Systems}, 35:19523--19536, 2022.

\bibitem[SGV24]{schlichtkrull2024averitec}
Michael Schlichtkrull, Zhijiang Guo, and Andreas Vlachos.
\newblock Averitec: A dataset for real-world claim verification with evidence from the web.
\newblock {\em Advances in Neural Information Processing Systems}, 36, 2024.

\bibitem[{\v{S}}PW23]{vsakota2023fly}
Marija {\v{S}}akota, Maxime Peyrard, and Robert West.
\newblock Fly-swat or cannon? cost-effective language model choice via meta-modeling.
\newblock {\em arXiv preprint arXiv:2308.06077}, 2023.

\bibitem[SYC{\etalchar{+}}20]{shridhar2020alfworld}
Mohit Shridhar, Xingdi Yuan, Marc-Alexandre C{\^o}t{\'e}, Yonatan Bisk, Adam Trischler, and Matthew Hausknecht.
\newblock Alfworld: Aligning text and embodied environments for interactive learning.
\newblock {\em arXiv preprint arXiv:2010.03768}, 2020.

\bibitem[TAB{\etalchar{+}}23]{team2023gemini}
Gemini Team, Rohan Anil, Sebastian Borgeaud, Yonghui Wu, Jean-Baptiste Alayrac, Jiahui Yu, Radu Soricut, Johan Schalkwyk, Andrew~M Dai, Anja Hauth, et~al.
\newblock Gemini: a family of highly capable multimodal models.
\newblock {\em arXiv preprint arXiv:2312.11805}, 2023.

\bibitem[TVCM18]{thorne2018fever}
James Thorne, Andreas Vlachos, Christos Christodoulopoulos, and Arpit Mittal.
\newblock Fever: a large-scale dataset for fact extraction and verification.
\newblock {\em arXiv preprint arXiv:1803.05355}, 2018.

\bibitem[TWL{\etalchar{+}}24]{trinh2024solving}
Trieu~H Trinh, Yuhuai Wu, Quoc~V Le, He~He, and Thang Luong.
\newblock Solving olympiad geometry without human demonstrations.
\newblock {\em Nature}, 625(7995):476--482, 2024.

\bibitem[WWS{\etalchar{+}}22]{wang2022self}
Xuezhi Wang, Jason Wei, Dale Schuurmans, Quoc Le, Ed~Chi, Sharan Narang, Aakanksha Chowdhery, and Denny Zhou.
\newblock Self-consistency improves chain of thought reasoning in language models.
\newblock {\em arXiv preprint arXiv:2203.11171}, 2022.

\bibitem[XCG{\etalchar{+}}23]{xi2023rise}
Zhiheng Xi, Wenxiang Chen, Xin Guo, Wei He, Yiwen Ding, Boyang Hong, Ming Zhang, Junzhe Wang, Senjie Jin, Enyu Zhou, et~al.
\newblock The rise and potential of large language model based agents: A survey.
\newblock {\em arXiv preprint arXiv:2309.07864}, 2023.

\bibitem[ZKAW23]{zhang2023ecoassistant}
Jieyu Zhang, Ranjay Krishna, Ahmed~H Awadallah, and Chi Wang.
\newblock Ecoassistant: Using llm assistant more affordably and accurately.
\newblock {\em arXiv preprint arXiv:2310.03046}, 2023.

\bibitem[ZKC{\etalchar{+}}24]{compound-ai-blog}
Matei Zaharia, Omar Khattab, Lingjiao Chen, Jared~Quincy Davis, Heather Miller, Chris Potts, James Zou, Michael Carbin, Jonathan Frankle, Naveen Rao, and Ali Ghodsi.
\newblock The shift from models to compound ai systems.
\newblock \url{https://bair.berkeley.edu/blog/2024/02/18/compound-ai-systems/}, 2024.

\end{thebibliography}
\bibliographystyle{alpha}
%\bibliographystyle{icml2024}
%\newpage
\appendix 
\section{Broader Impacts}\label{sec:scalelaw:impact}
Many inference strategies are often computationally and financially expensive to deploy due to multiple calls of language models. Our study suggests not blindly scaling up, but determining the resources allocated to these inference strategies carefully. More broadly, our paper paves the way for data-centric inference strategy designs. %In other words,   

\section{Limitations}\label{sec:scalelaw:limitation}
In this paper, we focus our analysis and experiments on the scaling behaviors for two instances of compound AI systems, but there are many other types of compound AI systems. For ease of evaluation, our experiments are conducted on relatively objective language tasks, but it remains open to study how the performance scales on subjective tasks such as generating poems and writing essays. Another interesting open problem is how to predict query difficult without querying the LMs.

\section{Missing Proofs}\label{sec:scalelaw:proofs}
We give all missing proofs here.
\subsection{Useful Lemmas}
\begin{lemma}\label{lemma:GPTNet:ExactFormula} Suppose $D$ is $2$-level difficult with $(\alpha, p_1,p_2)$ and  $\|A\|=2$, and W.L.O.G, $K$ is odd. Then 
$
    F(K;D) = \alpha I_{p_1}(\frac{K+1}{2},\frac{K+1}{2}) + (1-\alpha) I_{p_2}(\frac{K+1}{2},\frac{K+1}{2})
$, 
where $I_x(a,b)\triangleq \int_{0}^{x} t^{a-1} (1-t)^{b-1}dt/\int_{0}^{1} t^{a-1} (1-t)^{b-1}dt$ is the regularized incomplete beta function.
\end{lemma}
\begin{proof}
To analyze how the performance scales as the network size increases, let us first expand the performance of \LMNet{} by the law of total expectation
\begin{align}\label{eq:scalelaw:totallaw}
    F(K;D) = \mathbbm{E} [\hat{y}=y] = \sum_{i=1}^{2}\mathbbm{E} [\hat{y}=y|r(x)=p_i]\Pr[r(x)=p_i] 
\end{align}
Now consider $\mathbbm{E}[\hat{y}=y|r(x)=p]$ for any $p$. Note that there are in total 2 possible generations (by $\|A\|=2$), so the estimated generation $\hat{y}$ is correct if and only if more than half of the generations is correct. That is to say, $\mathbbm{E}[\hat{y}=y|r(x)=p] = \Pr[\sum_{k=1}^{K} \mathbbm{1}_{z_k=y}>\frac{K}{2}]  = 1 - \Pr[\sum_{k=1}^{K} \mathbbm{1}_{z_k=y}\leq\frac{K}{2}]$. For ease of notation, denote $Z\triangleq\sum_{k=1}^{K} \mathbbm{1}_{z_k=y}$. Note that $\mathbbm{1}_{z_k=y}$ is a Bernoulli variable with parameter $p$ and thus $Z$ is a Binomial variable with parameter $(K,p)$. Therefore, we have $Pr[Z\leq w] = I_{1-p}(K-w,w+1)$, where $I_x(a,b)\triangleq \int_{0}^{x} t^{a-1} (1-t)^{b-1}dt$ is the incomplete beta function. Thus, the above gives us
\begin{align*}
\mathbbm{E}[\hat{y}=y|r(x)=p] = 1 - I_{1-p}(K-\lfloor \frac{K}{2}\rfloor,\lfloor \frac{K}{2}\rfloor+1) = I_{p}(\lfloor \frac{K}{2}\rfloor+1, K-\lfloor \frac{K}{2}\rfloor) 
\end{align*}
If $K$ is odd, we can write it as 
$\mathbbm{E}[\hat{y}=y|r(x)=p]  = I_{p}(\frac{K+1}{2},\frac{K+1}{2}) 
$
We can now plug this back into equation (\ref{eq:scalelaw:totallaw}) to obtain 
\begin{align*}
    F(K;D) & = \sum_{i=1}^{2}\mathbbm{E} [\hat{y}=y|r(x)=p_i]\Pr[r(x)=p_i] =  \sum_{i=1}^{2}I_{p_i}(\frac{K+1}{2},\frac{K+1}{2})\Pr[r(x)=p_i] \\
    & = \alpha I_{p_1}(\frac{K+1}{2},\frac{K+1}{2}) + (1-\alpha)I_{p_2}(\frac{K+1}{2},\frac{K+1}{2})
\end{align*}
which completes the proof.
\end{proof}

\begin{lemma}\label{lemma:ScaleLaw:Beta}
    $I_p(M+1,M+1) = I_p(M,M) + 
        \frac{p^M (1-p)^M}{M B(M,M)} (2p-1)$, where $B(a,b)\triangleq \int_{0}^{1} t^{a-1}(1-t)^{b-1} d t$ is the beta function.
\end{lemma}
\begin{proof}
Noting that $I_p(a+1,b) = I_p(a,b)-\frac{p^a(1-p)^b}{a B(a,b)}$, we have  
\begin{align*}
    I_p(M+1,M+1) = I_p(M,M+1) - \frac{p^M(1-p)^{M+1}}{M B(M,M+1)} 
\end{align*}
By $I_p(a,b+1) = I_p(a,b)+\frac{p^a(1-p)^b}{b B(a,b)}$, we have
\begin{align*}
    I_p(M,M+1) = I_p(M,M) + \frac{p^M(1-p)^{M}}{M B(M,M)} 
\end{align*}
Thus, $ I_p(M+1,M+1)=\frac{p^M(1-p)^{M}}{M B(M,M)} - \frac{p^M(1-p)^{M+1}}{M B(M,M+1)}$. The Pascal's identity implies that $B(a,b+1) = \frac{a}{a+b} B(a,b)$ and thus $B(M,M+1) = B(M,M)/2$. Thus, we have
\begin{align*}
I_p(M+1,M+1)=\frac{p^M(1-p)^{M}}{M B(M,M)} - \frac{p^M(1-p)^{M+1}}{M B(M,M+1)} = \frac{p^M(1-p)^{M}}{M B(M,M)} - \frac{2 p^M(1-p)^{M+1}}{M B(M,M)}  
\end{align*}
Extracting the common factors gives 
\begin{align*}
\frac{p^M(1-p)^{M}}{M B(M,M)} - \frac{2 p^M(1-p)^{M+1}}{M B(M,M)}  =  \frac{p^M(1-p)^{M}}{M B(M,M)} [1-2(1-p)] = \frac{p^M(1-p)^{M}}{M B(M,M)} (2p-1) \\
\end{align*}
That is, $I_p(M+1,M+1) = I_p(M,M) + 
        \frac{p^M (1-p)^M}{M B(M,M)} (2p-1)$, which completes the proof.
\end{proof}

\subsection{Proof of Lemma \ref{lemma:scalelaw:difficultcondition}}
\begin{proof}
Let us first consider \VISFull{}. 

We want to first note that $\lim_{K\rightarrow\infty} \Pr[\hat{y}_K=y^*] = 1$, where  $y^* \triangleq \arg \max_{a\in A}\Pr[G(x,\theta)=a]$. To see this, we can first write $ \Pr[\hat{y}_K=y^*] = \Pr[\max_{a\not=y^*}\sum_{i=1}^{K} \mathbbm{1}_{z_i=a}<\sum_{i=1}^{K} \mathbbm{1}_{z_i=y^*}]=\Pr[\sum_{i=1}^{K} \mathbbm{1}_{z_i=a}-\sum_{i=1}^{K} \mathbbm{1}_{z_i=y^*}<0, \forall a\not=y^*]$. Subtracting 1 from both sides, we have $ 1- \Pr[\hat{y}_K=y^*] = \Pr[\cup_{a\not= y^*}\sum_{i=1}^{K} \mathbbm{1}_{z_i=a}-\sum_{i=1}^{K} \mathbbm{1}_{z_i=y^*}>0]$. 
Now by union bound, we can bound the right hand side by $\sum_{a\not=y} \Pr[\sum_{i=1}^{K} \mathbbm{1}_{z_i=a}- \mathbbm{1}_{z_i=y^*}>0] $. Now, note that each term within the inner summation is i.i.d. Furthermore, observe that the expectation is $\mathbbm{E}[\mathbbm{1}_{z_i=a}- \mathbbm{1}_{z_i=y^*}]= \Pr[G(x,\theta)=a] - \Pr[G(x,\theta)=y^*]<0$. Therefore, by  law of large numbers, we have $\lim_{K} \Pr[\sum_{i=1}^{K} \mathbbm{1}_{z_i=a}- \mathbbm{1}_{z_i=y^*}>0] = 0$. Since there are only finite number of $a\in A$, the sum of this quantity over all $a$ should also be 0. Therefore, we have  
\begin{equation*}
    \lim_{K\rightarrow \infty}1- \Pr[\hat{y}_K=y^*] = \lim_{K\rightarrow \infty}\Pr[\cup_{a\not= y^*}\sum_{i=1}^{K} \mathbbm{1}_{z_i=a}-\sum_{i=1}^{K} \mathbbm{1}_{z_i=y^*}>0] = 0
\end{equation*}
Thus, $\lim_{K\rightarrow \infty}\Pr[\hat{y}_K=y^*]=1$. 

Now, $d_V(x)>0$ implies $y=y^*$ and thus $\lim_{K\rightarrow\infty} F(K,D)=1$. $d_V(x)>0$ implies $y\not=y^*$ and thus $\lim_{K\rightarrow\infty} F(K,D)=0$.

Now let us consider \FVISFull{}. 

Abusing the notation a little, we claim that again $\lim_{K\rightarrow\infty} \Pr[\hat{y}_K=y^*] = 1$, where $y^* \triangleq \arg \max_{a\in A}\Pr[G_1(x,\theta)=a|\Phi(x,G_1(x,\theta),G_2(x,\theta))=1]$, where recall that $G_1(x,\theta), G_2(x,\theta)\triangleq G(x,\theta)$. To see this, we can first write $ \Pr[\hat{y}_K=y^*] = \Pr[\max_{a\not=y^*}\sum_{i=1}^{K} \mathbbm{1}_{z_i=a} w_i <\sum_{i=1}^{K} \mathbbm{1}_{z_i=y^*}w_i]=\Pr[\sum_{i=1}^{K} \mathbbm{1}_{z_i=a}w_i-\sum_{i=1}^{K} \mathbbm{1}_{z_i=y^*}w_i<0, \forall a\not=y^*]$. Subtracting 1 from both sides, we have $ 1- \Pr[\hat{y}_K=y^*] = \Pr[\cup_{a\not= y^*}\sum_{i=1}^{K} \mathbbm{1}_{z_i=a}w_i-\sum_{i=1}^{K} \mathbbm{1}_{z_i=y^*}w_i>0]$. 
Now by union bound, we can bound the right hand side by $\sum_{a\not=y} \Pr[\sum_{i=1}^{K} \mathbbm{1}_{z_i=a}w_i- \mathbbm{1}_{z_i=y^*}w_i>0] $. Now, note that each term within the inner summation is i.i.d. Furthermore, observe that the expectation is $\mathbbm{E}[\mathbbm{1}_{z_i=a}w_i- \mathbbm{1}_{z_i=y^*}w_i]= \Pr[G_1(x,\theta)=a, \Phi(x,G_1(x,\theta),G_2(x,\theta))=1] - \Pr[G(x,\theta)=y^*,\Phi(x,G_1(x,\theta),G_2(x,\theta))=1]<0$. Therefore, by  law of large numbers, we have $\lim_{K} \Pr[\sum_{i=1}^{K} \mathbbm{1}_{z_i=a} w_i- \mathbbm{1}_{z_i=y^*}w_i>0] = 0$. Since there are only finite number of $a\in A$, the sum of this quantity over all $a$ should also be 0. Therefore, we have  
\begin{equation*}
    \lim_{K\rightarrow \infty}1- \Pr[\hat{y}_K=y^*] = \lim_{K\rightarrow \infty}\Pr[\cup_{a\not= y^*}\sum_{i=1}^{K} \mathbbm{1}_{z_i=a}w_i- \mathbbm{1}_{z_i=y^*} w_i>0] = 0
\end{equation*}
Thus, $\lim_{K\rightarrow \infty}\Pr[\hat{y}_K=y^*]=1$. 

Now, $d_F(x)>0$ implies $y=y^*$ and thus $\lim_{K\rightarrow\infty} F(K,D)=1$. $d_F(x)>0$ implies $y\not=y^*$ and thus $\lim_{K\rightarrow\infty} F(K,D)=0$. 

Hence, we have shown that $d_V(x)$ and $d_F(x)$ are difficulty indicator for \VISFull{} and \FVISFull{}, respectively. 
\end{proof}

\subsection{Proof of Theorem \ref{thm:scaleLaw:character}}
\begin{proof}
We construct a recurrent relation on $F(K;D)$ by Applying Lemma \ref{lemma:GPTNet:ExactFormula} 
\begin{align*}
&F(K+2;D) - F(K;D)\\ = & \alpha I_{p_1}(\frac{K+3}{2},\frac{K+3}{2}) + (1-\alpha) I_{p_2}(\frac{K+3}{2},\frac{K+3}{2}) - [ \alpha I_{p_1}(\frac{K+1}{2},\frac{K+1}{2}) + (1-\alpha) I_{p_2}(\frac{K+1}{2},\frac{K+1}{2})]\\ = & \alpha( I_{p_1}(\frac{K+3}{2},\frac{K+3}{2}) - I_{p_1}(\frac{K+1}{2},\frac{K+1}{2})) + (1-\alpha)  (I_{p_2}(\frac{K+3}{2},\frac{K+3}{2}) - I_{p_2}(\frac{K+1}{2},\frac{K+1}{2})) 
\end{align*}
For ease of notation, let us denote $M=\frac{K+1}{2}$. Applying Lemma \ref{lemma:ScaleLaw:Beta} leads to 
\begin{align*}
&F(K+2;D) - F(K;D)\\
= & \alpha [ \frac{p_1^M (1-p_1)^M}{M B(M,M)} (2p_1-1) ] +  (1-\alpha) [ \frac{p_2^M (1-p_2)^M}{M B(M,M)} (2p_2-1) ]\\
= & \frac{1}{MB(M,M)} [ \alpha \cdot p_1^M (1-p_1)^M (2p_1-1) + (1-\alpha) \cdot p_2^M (1-p_2)^M (2p_2-1) ] 
\end{align*}
Now rearranging the terms gives
\begin{align*}
&F(K+2;D) - F(K;D)\\
= & \frac{1}{MB(M,M)}\cdot (1-\alpha) p_2^M (1-p_2)^M (1-2p_2) \cdot [ \frac{\alpha (2p_1-1)}{(1-\alpha)(1-2p_2) } [\frac{p_1(1-p_1)}{p_2(1-p_2)}]^M - 1 ]  
\end{align*}
For ease of notation, denote $\Delta F(M) \triangleq \frac{\alpha (2p_1-1)}{(1-\alpha)(1-2p_2) } 
 \frac{p_1(1-p_1)}{p_2(1-p_2)}]^M - 1$. 
Note that $\Delta F(M)$ is monotonically increasing or decreasing, depending on the parameters $p_1, p_2$. It is also easy to show that $\alpha \geq 1-\frac{1}{t}$ if and only if $\Delta F(1) \geq 0$. Now, we are ready to derive the main results by analyzing $\Delta F(M)$.

\begin{itemize}
    \item $p_1+p_2>1$ and $\alpha \geq 1-\frac{1}{t}$: Now $\frac{p_1(1-p_1)}{p_2(1-p_2)}>1$, and thus $\Delta F(M)$ is monotonically increasing. Furthermore, it is easy to see $\lim_{M \rightarrow \infty} \Delta F(M)= \infty$. Furthermore,  $\Delta F(1) \geq 0$. That is, for any given $M$, $\Delta F(M)$ is non-negative and thus $F(K;D)$ must be monotonically increasing
    \item $p_1+p_2>1$ and $\alpha <  1-\frac{1}{t}$: Now $\frac{p_1(1-p_1)}{p_2(1-p_2)}>1$, and thus $\Delta F(M)$ is monotonically increasing. Furthermore, it is easy to see $\lim_{M \rightarrow \infty} \Delta F(M)= \infty$. Furthermore,  $\Delta F(1) < 0$. That is, as $K$ and thus $M$ increases,  $\Delta F(M)$ is negative and then becomes positive. Therefore, $F(K;D)$ must decrease and then increase
    \item $p_1+p_2<1$ and $\alpha > 1-\frac{1}{t}$: Now $\frac{p_1(1-p_1)}{p_2(1-p_2)}<1$, and thus $\Delta F(M)$ is monotonically decreasing. Furthermore, it is easy to see $\lim_{M \rightarrow \infty} \Delta F(M)= -1$. Furthermore,  $\Delta F(1) > 0$. That is, as $K$ and thus $M$ increases,  $\Delta F(M)$ is positive and then becomes negative. Therefore, $F(K;D)$ must increase and then decrease
    
    \item $p_1+p_2<1$ and $\alpha \leq 1-\frac{1}{t}$: Now $\frac{p_1(1-p_1)}{p_2(1-p_2)}<1$, and thus $\Delta F(M)$ is monotonically decreasing. Furthermore, it is easy to see $\lim_{M \rightarrow \infty} \Delta F(M)= -1$. Furthermore,  $\Delta F(1) \leq 0$. That is, for any given $M$, $\Delta F(M)$ is non-positive and thus $F(K;D)$ must be monotonically decreasing
\end{itemize}
Thus, we complete the proof. 
\end{proof}

\subsection{Proof of Theorem \ref{thm:scalelaw:scalingmodel}}
\begin{proof}
Let us first consider $F(K,x)=\Pr[\hat{y}_K=y]$.  Note that there are in total 2 possible generations (by $\|A\|=2$), so the estimated generation $\hat{y}$ is correct if and only if more than half of the generations is correct. That is to say, $\mathbbm{E}[\hat{y}_K=y] = \Pr[\sum_{k=1}^{K} \mathbbm{1}_{z_k=y}>\frac{K}{2}]  = 1 - \Pr[\sum_{k=1}^{K} \mathbbm{1}_{z_k=y}\leq\frac{K}{2}]$. For ease of notation, denote $Z\triangleq\sum_{k=1}^{K} \mathbbm{1}_{z_k=y}$. Note that $\mathbbm{1}_{z_k=y}$ is a Bernoulli variable with parameter $p$ and thus $Z$ is a Binomial variable with parameter $(K,p)$. Therefore, we have $Pr[Z\leq w] = I_{1-p}(K-w,w+1)$, where $I_x(a,b)\triangleq \int_{0}^{x} t^{a-1} (1-t)^{b-1}dt/\int_{0}^{1} t^{a-1} (1-t)^{b-1}dt$ is the incomplete beta function. Specifically, we can set $w=\frac{K}{2}$ and get $F(K,x)=\Pr[\hat{y}_K=y]=I_{1-p}(\frac{K+1}{2},\frac{K+1}{2})$. Noting that by definition, $d_V(x)=1-2p$, we can rewrite this as  $F(K,x)=I_{\frac{1-d_V(x)}{2}}(\frac{K+1}{2},\frac{K+1}{2})$. Taking the integral over $x$ completes the proof.
\end{proof}

\subsection{Proof of Theorem \ref{thm:scalelaw:optimalnetworksize}}
\begin{proof}
   Note that the optimal number of LM calls must ensure that the incremental component $\Delta F(M)=0$.  That is, the optimal value $M^*$ must satisfy 
$
    \Delta F(M^*) = \frac{\alpha (2p_1-1)}{(1-\alpha)(1-2p_2) } 
 [\frac{p_1(1-p_1)}{p_2(1-p_2)}]^{M^*} - 1=0.  
  $
   Solving the above equation gives us a unique solution 
   \begin{align*}
       M^*= 2 \frac{\log\frac{\alpha}{1-\alpha}\frac{2p_1-1}{1-2p_2}}{\log \frac{p_2(1-p_2)}{p_1(1-p_1)}} 
   \end{align*}
Noting that $K^*=\lceil 2M \rceil]$ completes the proof.
\end{proof}

\section{Additional Experiment Details}\label{sec:scalelaw:exp_appendix}

\subsection{Datasets}
We evaluate \VISFull{} and \FVISFull{} on four real-world datasets, namely, GPQA~\cite{rein2023gpqa}, AVERITEC~\cite{schlichtkrull2024averitec}, TRUTHFULQA~\cite{lin2021truthfulqa}, and MMLU PHYSICS~\cite{hendrycks2020measuring}.   
AVERITEC is a fact verification dataset, where each query is a claim (e.g., ``All USA ballots Arriving after election day will be thrown out''), and the goal is to determine if this claim should be refused, supported, or there is conflicting evidence or not enough confidence.  We use the ``development'' partition offered in the original paper, which contains 500 queries. 
GPQA is a dataset consisting of multiple-choice questions written by domain experts in biology, physics, and chemistry. We use the diamond partition, which contains 198 queries, and as reported by the original paper, enjoys the highest expert accuracy. 
TRUTHFULQA is another widely used multiple-choice question-answering dataset. We uses the validation parititon, which contains 817 questions spanning 38 categories, including health, law, finance, and politics. MMLU PHYSICS is the high school physics subset of the MMLU dataset, which contains 151 questions. All queries are prompted as multiple-choice questions for ease of evaluation.

\paragraph{Liscense.} GPQA is released under the MIT License.  AVERITEC uses  Creative Commons Attribution-NonCommercial 4.0 International License. TRUSTFULQA and PHYSICS (as part of MMLU) both adopt the Apache License 2.0

\subsection{Prompting for generators and filters}
We provide the prompts used for the generators and filters in the following boxes.

\begin{tcolorbox}[colback=gray!5!white,colframe=black!75!black,title=Prompt for the generator $G()$]
Please answer the following question. You should first analyze it step by step.  Then generate your final answer by the answer is (X).

Q: \{Query\}

A: 
\end{tcolorbox}

\begin{tcolorbox}[colback=gray!5!white,colframe=black!75!black,title=Prompt for the filter $\Phi()$]
[User Question]: \{query\}

[Answer]:\{answer\}

Instruction: Review your previous answer and find problems with your answer. Finally, conclude with either ’[[correct]]’ if the above answer is correct or ’[[wrong]]’ if it is incorrect. Think step by step.

Verdict:
\end{tcolorbox}

\subsection{Generation Setup}
We set up the temperature as 0.1 since all the evaluation tasks are objective and have clear correct answers. For each question, we query GPT-3.5-turbo-0125 400 times. Then for each $F(K,D)$, we randomly sample $K$ answers from the 400 answers with replace to simulate the performance once, and report the average over 1000 runs.

\subsection{The optimal number of LM calls predicted by the scaling model}

\begin{table}[t]
  \centering
  \small
  \caption{Optimal number of LM calls prediction. For each data distribution with specific 2-level difficulty parameters $\alpha, p_1, p_2$, we sample 100 data points, employ a simulated Voting \LMNet{} with 1000 number of LM calls, estimate the parameters by their empirical mean and adopt the analytical optimal number of LM calls predictor.  Across evaluated \dataitem{} difficulties, our predicted optimal number of LM calls exactly matches the optimal number empirically observed.}
    \begin{tabular}{||c|c|c|c|c|c|c||}
    \hline
     \multirow{2}[4]{*}{$\alpha$} &
      \multirow{2}[4]{*}{$p_1$} &
      \multirow{2}[4]{*}{$p_2$} &
      \multirow{2}[4]{*}{$\hat{\alpha}$} &
      \multirow{2}[4]{*}{$\hat{p}_1$} &
      \multirow{2}[4]{*}{$\hat{p}_2$} &
      
      \multicolumn{1}{l|}{Optimal Number of LM Calls}
      \bigstrut\\
\cline{7-7}     &
       &
       &
       &
       &
       &
      Analytical/Empirical
      \bigstrut\\
    \hline
    \hline
    0.4 &
      0.85 &
      0.4 &
      0.42  &
      0.84  &
      0.40  &
      3
      \bigstrut\\
    \hline
    0.4 &
      0.85 &
      0.3 &
      0.38  &
      0.85  &
      0.30  &
      1
      \bigstrut\\
    \hline
    0.4 &
      0.75 &
      0.4 &
      0.41  &
      0.74  &
      0.40  &
      4
      \bigstrut\\
    \hline
    0.4 &
      0.75 &
      0.3 &
      0.44  &
      0.75  &
      0.30  &
      1
      \bigstrut\\
    \hline
    0.4 &
      0.65 &
      0.4 &
      0.41  &
      0.65  &
      0.40  &
      1
      \bigstrut\\
    \hline
    0.5 &
      0.85 &
      0.4 &
      0.50  &
      0.84  &
      0.40  &
      4
      \bigstrut\\
    \hline
    0.5 &
      0.85 &
      0.3 &
      0.50  &
      0.85  &
      0.30  &
      2
      \bigstrut\\
    \hline
    0.5 &
      0.75 &
      0.4 &
      0.51  &
      0.75  &
      0.40  &
      7
      \bigstrut\\
    \hline
    0.5 &
      0.75 &
      0.3 &
      0.49  &
      0.75  &
      0.30  &
      4
      \bigstrut\\
    \hline
    0.5 &
      0.65 &
      0.4 &
      0.48  &
      0.65  &
      0.40  &
      15
      \bigstrut\\
    \hline
    0.6 &
      0.85 &
      0.4 &
      0.59  &
      0.84  &
      0.39  &
      5
      \bigstrut\\
    \hline
    0.6 &
      0.85 &
      0.3 &
      0.63  &
      0.85  &
      0.30  &
      4
      \bigstrut\\
    \hline
    0.6 &
      0.75 &
      0.4 &
      0.61  &
      0.75  &
      0.40  &
      11
      \bigstrut\\
    \hline
    0.6 &
      0.75 &
      0.3 &
      0.59  &
      0.75  &
      0.30  &
      11
      \bigstrut\\
    \hline
    0.6 &
      0.65 &
      0.4 &
      0.62  &
      0.65  &
      0.40  &
      30
      \bigstrut\\
    \hline
    \end{tabular}%
  \label{tab:ScaleLaw:OptimalNetworkSize}%
\end{table}%

Table \ref{tab:ScaleLaw:OptimalNetworkSize} compares the optimal number of LM calls predicted by the scaling model and the ground-truth value on the bi-level dataset $D_{\alpha,p_1,p_2}$. Overall, We observe that they match very well.

\subsection{Performance Estimation via Our Proposed Scaling Law}
\begin{figure}[t]
    \centering
\begin{subfigure}[t]{0.33\textwidth}
        \centering
\includegraphics[width=0.99\textwidth]{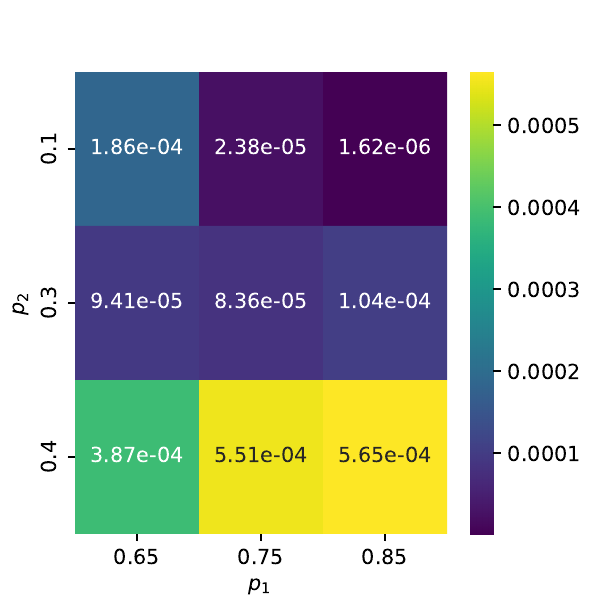}
        \caption{$\alpha=0.4$}
\end{subfigure}%\label{fig:enter-label}
\begin{subfigure}[t]{0.33\textwidth}
        \centering
\includegraphics[width=0.99\textwidth]{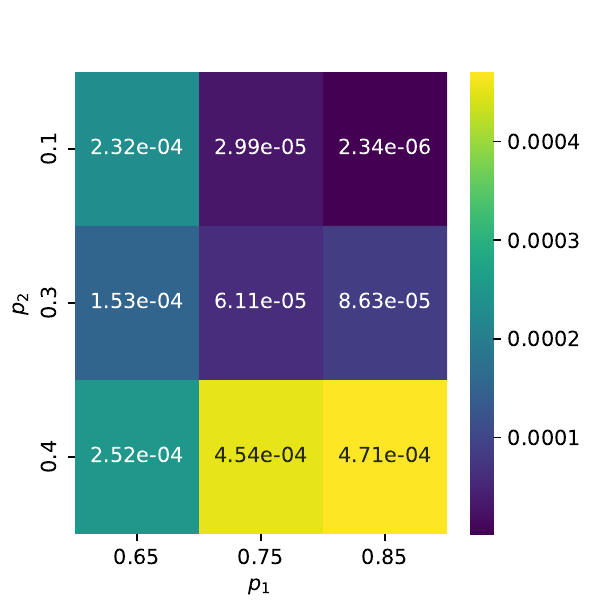}
        \caption{$\alpha=0.5$}
\end{subfigure}%\label{fig:enter-label}
\begin{subfigure}[t]{0.33\textwidth}
        \centering
\includegraphics[width=0.99\textwidth]{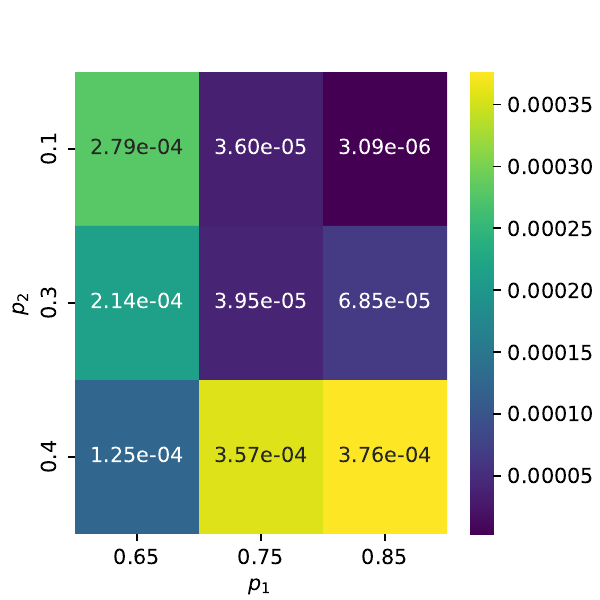}
        \caption{$\alpha=0.6$}
\end{subfigure}%
\caption{Mean square error of the performance predicted by our proposed scaling law on synthesized datasets with varying bi-level difficulties. Here, we fit the scaling law by performance evaluated at $K=1,2,3,4,5$, and evaluate its performance for $K$ from 1 to 100.  Overall, we observe that the predicted performance accurately matches the empirical evaluation.}\label{fig:ScaleLaw:Error}
\end{figure}
Can our proposed scaling model predict the empirical performance of \VISFull{} accurately?  To answer this, we apply it to each dataset considered in Figure \ref{fig:ScaleLaw:Simulation}. In particular, we feed our estimator with the \LMNets{}' performance with the number of LM calls being $1, 2, 3, 4, 5$, and then use it to predict the performance for LM calls within the range from 1 to 100. Note that this is quite challenging, as the estimator needs to extrapolate, i.e.,  predict the performance of a Voting \LMNet{} whose number of LM calls is much larger than any number seen in the training data. 

As shown in Figure \ref{fig:ScaleLaw:Error}, the performance predicted by our estimator accurately matches the empirical observation. Across all \dataitem{} difficulty parameters, the mean square error ranges from $1e-6$ to $1e-4$. This suggests that predicting how the number of LM calls affects a Voting \LMNet{} is feasible and also indicates that our scaling law captures the key performance trend effectively.

\end{document}